\documentclass[10pt,twocolumn,letterpaper]{article}

\usepackage{cvpr}
\usepackage{times}
\usepackage{epsfig}

\usepackage{graphicx,subfigure}
\usepackage{multirow}
\usepackage{amsmath}
\usepackage{amssymb}
\usepackage{color}
\usepackage{enumerate}
\graphicspath{{./fig/}}
\usepackage{floatrow}
\usepackage{wrapfig}
\newfloatcommand{capbtabbox}{table}[][\FBwidth]

\newfloatcommand{capbfigbox}{figure}[][\FBwidth]

\cvprfinalcopy 

\def\cvprPaperID{4} 

\newcommand{\ea}{et al. }
\begin{document}

\title{What Will I Do Next? \\The Intention from Motion Experiment}

\author{Andrea Zunino$^{\pmb{1},\pmb{3}}$, Jacopo Cavazza$^{\pmb{1},\pmb{3}}$, Atesh Koul$^{\pmb{2}}$, Andrea Cavallo$^{\pmb{2},\pmb{4}}$, \\ Cristina Becchio$^{\pmb{2},\pmb{4}}$ and Vittorio Murino$^{ \pmb{1},\pmb{5}}$\\ 
	{\tt \small firstname.lastname@iit.it} \\
	$^{\pmb 1}$ Pattern Analysis \& Computer Vision (PAVIS) -- Istituto Italiano di Tecnologia -- \textit{Genova, Italy} \\
	$^{\pmb 2}$ Robotics, Brain and Cognitive Science -- Istituto Italiano di Tecnologia (IIT) -- \textit{Genova, Italy} \\
	$^{\pmb 3}$ Electrical, Electronics and Telecommunication Engineering and Naval Architecture Department \\ (DITEN) -- Universit\`{a} degli Studi di Genova --  \textit{Genova, Italy}\\
	$^{\pmb 4}$ Psychology Department -- University of Torino -- \textit{Torino, Italy} \\
	$^{\pmb 5}$Computer Science Department -- Universit\`{a} di Verona --  \textit{Verona, Italy}}

\maketitle
\thispagestyle{empty}

\begin{abstract}
	In computer vision, video-based approaches have been widely explored for the early classification and the prediction of actions or activities. However, it remains unclear whether this modality (as compared to 3D kinematics) can still be reliable for the prediction of human intentions, defined as the overarching goal embedded in an action sequence. Since the same action can be performed with different intentions, this problem is more challenging but yet affordable as proved by quantitative cognitive studies which exploit the 3D kinematics acquired through motion capture systems. 
	
	In this paper, we bridge cognitive and computer vision studies, by demonstrating the effectiveness of video-based approaches for the prediction of human intentions. Precisely, we propose {\em Intention from Motion}, a new paradigm where, without using any contextual information, we consider instantaneous grasping motor acts involving a bottle in order to forecast why the bottle itself has been reached (to pass it or to place in a box, or to pour or to drink the liquid inside).
	
	We process only the grasping onsets casting intention prediction as a classification framework. Leveraging on our multimodal acquisition (3D motion capture data and 2D optical videos), we compare the most commonly used 3D descriptors from cognitive studies with state-of-the-art video-based techniques. Since the two analyses achieve an equivalent performance, we demonstrate that computer vision tools are effective in capturing the kinematics and facing the cognitive problem of human intention prediction.
\end{abstract}

\section{Introduction}
\begin{figure*}[t!]
	\centering
	\subfigure[Action/activity recognition\label{s:AR}]{\includegraphics[height=0.12\textheight,width=.4\textwidth]{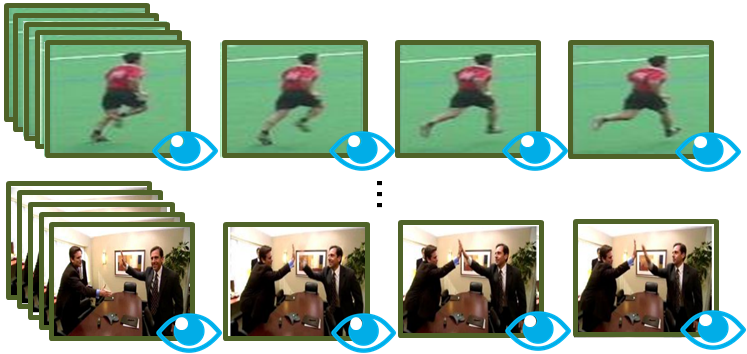}}\qquad\quad
	\subfigure[Early activity recognition (EAR) \label{s:EAR}]{\includegraphics[height=0.12\textheight,width=.4\textwidth]{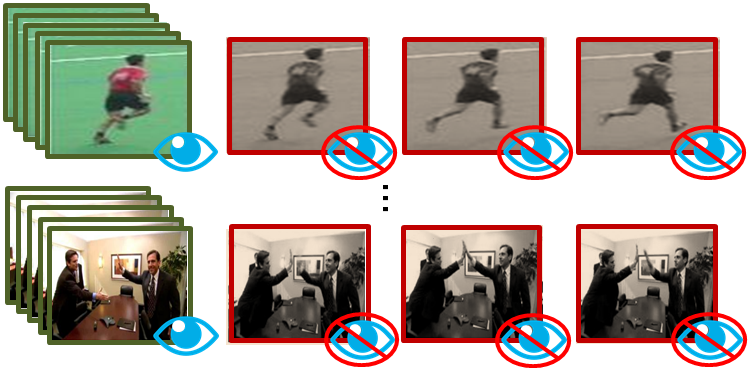}}\\
	\subfigure[Action prediction (AP) \label{s:AP}]{\includegraphics[height=0.2\textheight,width=.46\textwidth]{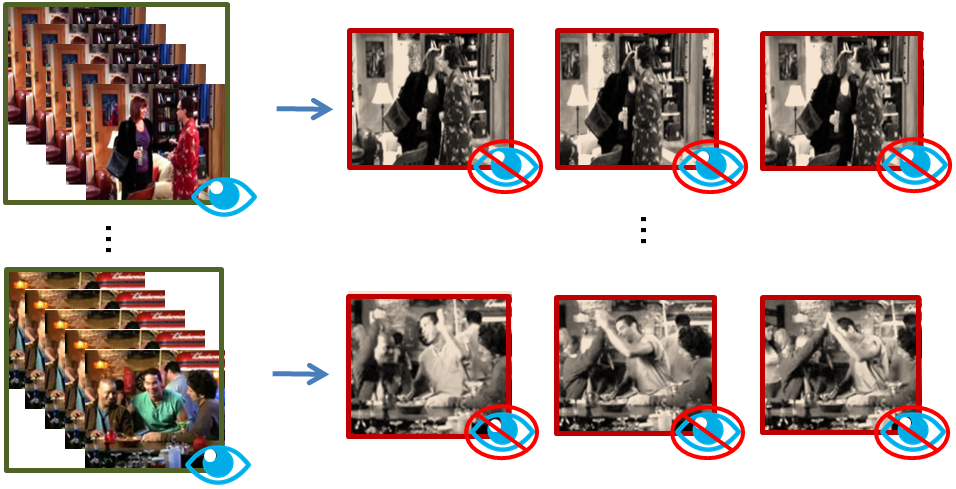}}\qquad\quad
	\subfigure[{\bf Intention prediction} \label{s:IfM} ]{\includegraphics[height=0.2\textheight,width=.46\textwidth]{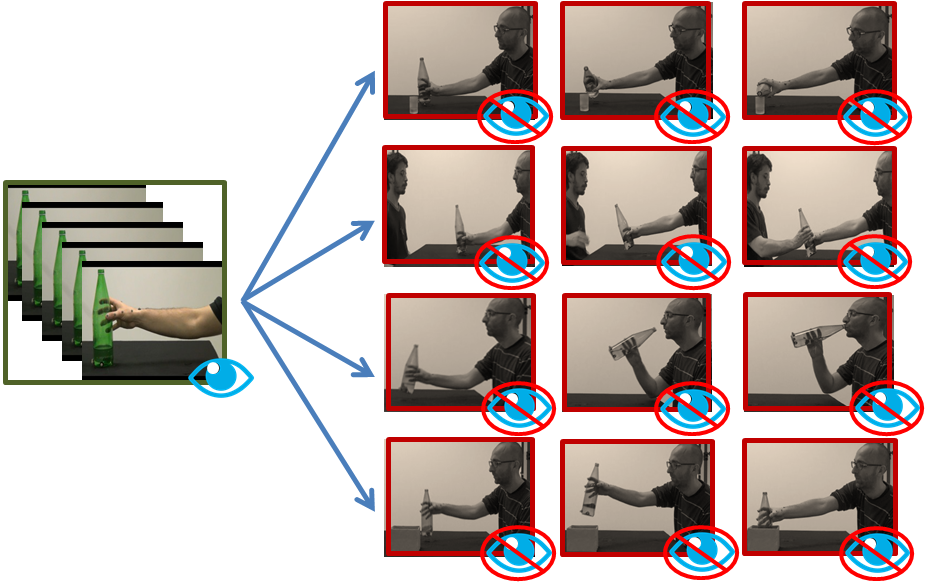}}
	\caption{Four different paradigms in human action/activity analysis. \ref{s:AR} Action/activity recognition: each sequence is fully observed to infer the class label (``running'' for the top sequence up to ``high-five'' for the bottom). \ref{s:EAR} Early activity recognition: only a few initial frames per sequence is observed and the goal is an early classification from these incomplete observations. \ref{s:AP} Action prediction: future actions are predicted analysing all past events which, in general, can be very different across different classes. Thus, in the top sequence a standing up activity leads to predict a ``kissing'', while, in the bottom, a conversation between a group of friends anticipates a ``high-five''. \ref{s:IfM} Intention prediction: a novel paradigm where unobserved future action are anticipated from the same class of motor act, all extremely similar in appearance, no matter which different ending will occur.} 
	\label{fig:ciao}
\end{figure*}

Action and activity recognition are one of the most active areas in computer vision. The task here consists in the classification of \emph{fully observed} sequence and many methods have been proposed to tackle this task \cite{Laptev:2005,Wang:2011}. More recently, the community has also started to investigate a few variants, either performing the early classification of partially disclosed activities or predicting future actions by analyzing the events occurring up to a certain instant. For the sake of clarity, let us briefly review these two different paradigms which are also sketched in Fig. \ref{fig:ciao}. \\

{\bf Early Activity Recognition (EAR).} Ryoo \cite{Ryoo} devised a system to infer the ongoing activity by only analysing its \emph{onset}, \ie the initial part of the action. This is done with a dynamic programming method to match an extension of classical bag-of-features representation which allow to capture the temporal correlation of descriptors. 
Hoai and De la Torre \cite{Hoai} designed MMED, Max-Margin Early-event Detectors to address the problem to understand a specific human emotion after it starts but before it ends. They trained a structured output support vector machine \cite{SOSVM:2005} on the whole accomplished emotion development and, during testing phase, they managed to classify fear rather than disgust when they are about to occur. Yu \ea   \cite{Yu:ACM12} propose a generalization of Spatio-Temporal Interest Point \cite{Laptev:2005} to categorize actions from their beginning. The inter-dependencies between different spatial location are implemented into a probabilistic graphical model fed by histogram features. Cao \ea  \cite{Cao:CVPR13} split a complete action into temporal segments which are further represented by means of sparse coding, so that actions are recognizable from incomplete data. Ryoo \ea  \cite{Ryoo:HRI15} tackle early activity recognition from egocentric videos: the task is detecting the so-called \emph{onset signature}, a bunch of kinematic evidence which has strong predictive properties about the last part of the observed action.  Some works have attempted to investigate how much of the whole action is necessary to perform a classification: Davis and Tyagi \cite{Davis:imavis06} adopt a generative probabilistic framework to deal with the uncertainty due to limited amount of data, while Schindler and Van Gool \cite{Schindler:CVPR08} try to answer the aforementioned question using a similarity measure between the statical and the motion information extracted from videos.
\\
{\bf Action Prediction (AP).} Lan and Savarese \cite{Savarese14} developed the so-called \emph{hierarchical movemes}, a new representation to model human actions from coarse to fine levels of granularities which was integrated in a max-margin learning framework for action prediction.
Vondrick and Torralba \cite{Torralba:15} uses a deep neural network trained over 600 hours of videos. During training the net exploits videos to learn to predict the representation of frames in the future and the last fully connected layer allows to perform classification over different future endings. For the sake of prediction, contextual information plays a pivotal role. Indeed, once the objects present in a scene are detected, the object-object or object-person relationship can be modelled by several probabilistic architectures (\eg, graphical models \cite{Amit:ACCV14,FZ:CVPR14} or topic models \cite{Koppula,Pei:2011}). Among the works which directly models the context inside the algorithms, some of them deal with the prediction of future trajectories of moving objects (vehicles or pedestrian) \cite{Kitani,Walker,YamaguchiCVPR11,Xie:2013} by estimating the spatial areas over which such objects will most likely pass with respect to those which are excluded by this passage (\eg, car circulations over sidewalks \cite{Walker}).

{\bf Shortcomings in EAR and AP.} Previous attempts in EAR and AP frequently exploit motion patterns which are specific of the subsequent actions, since they contain some cues that undoubtedly help the recognition. 
For instance, if the goal is understanding whether two people are going to shake their hands or to give a high-five, by just looking at the first part of their interaction, a low wrist height can be an evidence of a handshaking \cite{Torralba:15,Savarese14}. 
Further, another important aspect of the entire activity recognition problem is that the current techniques are mainly exploiting the scene context to support the classification (\cite{Cao:CVPR13,FZ:CVPR14,Amit:ACCV14,Xie:2013,Walker,Pei:2011,Amit:ACCV14,Kilner:2011,vanElk:2014}, and \cite{Bubetal:2013}), i.e., the objects present in the scene and the knowledge about the actions associated to them are cues that can be utilized to help in making a correct inference of the ongoing action to be recognized. This information is necessary but often insufficient to solve the issue or, worse, the context may not always be available or easily recognizable, being also misleading when the scene is too noisy or cluttered \cite{ZeB:15}. In any case, an important source of information to disambiguate intention can be provided by the kinematics of the movement.

{\bf Cognitive studies.} Recent findings in cognition indicate that how a motor act is performed (e.g., grasping an object) is not solely determined by biomechanical constraints imposed by the object extrinsic and intrinsic properties with which one is interacting, but depends on the agent’s intention.
Indeed, intentions become "visible" in the displayed actions of agents manipulating a given object \cite{Asuini:2014,Asuini:2015}, and this process is actually readable by observers. Manera et al. \cite{Manera:2011} showed that in a binary choice design, observers were able to judge whether the agent’s intent in grasping the object was to cooperate with a partner or compete against an opponent.  In addition, by means of a motion capture system to acquire the exact kinematics, Ansuini et al. \cite{Asuini:2014} proved that how a given object is grasped is richly informative for observer to determine the social vs. individual agent's intention. Recently, Becchio et al. \cite{Becchio_ScientReport} demonstrated that it is possible to quantify the impact of kinematic variables (such as wrist height/velocity) for the sake of intentions' recognition. \\
The analysis is extremely interesting since may corroborate the actual feasibility of predicting intentions even in context-free settings, where kinematics is the only available source of information. Nevertheless, there is a gap between quantitative evidences drawn from 3D kinematics \cite{Asuini:2014,Becchio_ScientReport} and human performance analysis carried out on video data \cite{Manera:2011}. Specifically, it is still unclear whether the subtle kinematics which differentiate intentions is also exploitable from RGB videos in a quantitative fashion.

\subsection{Paper Contributions}

In this paper, we aim at introducing {\bf Intention from Motion} (IfM), a brand new problem with two challenging aspects largely differentiate this work from the current literature of either EAR or AP. 

\begin{itemize}	
	\item Grounding from the assumption that the same class of motor acts can be performed with different intentions, we want to analyse the movement onset of an apparently unrelated action (actually embedding the intention from the very beginning), the same for all intentions, capturing those subtle motion patterns which are anticipative of the future action. 
	\item Unlike the main existing literature, we want to avoid the exploitation of any hints derived by the context, solely focusing on the kinematics of the movement. 
\end{itemize}


This novel setup was accomplished by a set of experiments where subjects were asked to grasp a bottle, in order to either 1) pour some water into a glass, 2) pass the bottle to a co-experimenter, 3) drink from it, or 4) place the bottle into a box. The dataset is composed by both 3D trajectories of twenty motion capture (VICON) markers outfitted over the hand of the participants and optical RGB video sequences lasting about one second, with an occlusive camera view in which only the arm and the bottle are visible. Data are acquired from the moment when the hand starts from a stable fixed position up to the reaching of the object, and 3D marker trajectories and video sequences are exactly trimmed at the instant when the hand grasps the bottle, removing the following part. The goal is to classify the intentions associated with the observed grasping-a-bottle movement, \ie to predict the agent's intention.

Even if some methodologies have been proposed to for EAR and AP \cite{Savarese14,Ryoo,Torralba:15,Hoai,Walker}, the experiments are typically performed on standard action recognition datasets, just adapted to the new task, and often considering the start of the same action which of course helps. On the contrary, our experiment is explicitly designed for intention prediction.

Due to the multimodality of our experimental settings, our work interleaves the EAR and AP problems from computer vision with the kinematic approach of cognitive studies in predicting intentions. Indeed, by taking advantage of state-of-the-art hand-crafted features in video-based action recognition, we are able to prove that the subtle kinematic analysis required to predict intentions is actually feasible while leveraging on RGB optical video as an alternative approach to motion capture data \cite{Asuini:2014,Becchio_ScientReport} which are obviously more difficult to obtain.

In summary, the present work introduces the following main contributions.



\begin{enumerate}[$(a)$]
	
	\item We introduce the new problem of Intention from Motion. That is, from the same observable ``neutral'' motor act - used in both training and test phases - we try to classify the underlying intention using solely motion information, without any contextual cue. Unlike previous works, we are neither classifying actions from their very first beginning (e.g., \cite{Ryoo,Hoai}) nor classifying different futures by analyzing different past onsets \cite{Savarese14,Torralba:15}. Instead, we anticipate intentions which finalize the same class of motor act, distilling from it the discriminative motion patterns characterizing the specific intention while fully neglecting any contextual information (as opposed to \cite{Amit:ACCV14} or \cite{Walker}).
	
	\item  We propose a (3D + 2D) dataset specifically aimed at the prediction of human intentions. This dataset is designed in a principled way by defining four intentions (Pouring, Passing, Drinking, Placing) performed by independent naive subjects, which are all forerun from the very similar initial movement of grasping-a-bottle, while avoiding bias which can affect the subsequent performance analysis. To the best of our knowledge, this is the first time a dataset has been explicitly designed for intention prediction.
	
	\item We carry out a separate 3D and 2D analysis, exploiting either broadly used 3D kinematic features or state-of-the-art video-based approaches. In light of the equivalent performance obtained by the two, we bridge the gap between video-based approaches for early activity recognition/action prediction (computer vision) and the analysis of 3D kinematics acquired by motion capture (cognitive studies). 
\end{enumerate}


\paragraph{Paper outline.} Section \ref{sez:dataset} introduces our dataset and the experimental setting. We investigate human performance in intention prediction in Section \ref{sez:uomo}. Subsequently, we extensively describe the experimental analysis accomplished on the 3D data in Section \ref{sez:3D}, and on the 2D video sequences in Section \ref{sez:2D}. Section \ref{sez:con} finally draws the conclusions.





\section{Dataset overview}\label{sez:dataset}

Seventeen naive volunteers were seated beside a $110 \times 100$ cm table resting on it elbow, wrist and hand inside a tape-marked starting point. A glass bottle was positioned on the table at a distance of about $46$ cm and participants were asked to grasp it in order to perform one of the following 4 different intentions.
\begin{enumerate}
	\vspace{-.3 cm}
	\item{ \bf Pouring} some water into a small glass (diameter $5$ cm; height $8.5$ cm) positioned on the left side of the bottle, at $25$ cm from it.\vspace{-.3 cm}
	\item{\bf Passing} the bottle to a co-experimenter seating opposite the table.\vspace{-.3 cm}
	\item{\bf Drinking} some water from the bottle.\vspace{-.25 cm}
	\item {\bf Placing} the bottle in a cardboard $17 \times 17 \times 12.5$ box positioned on the same table, $25$ cm distant.\vspace{-.2 cm}
\end{enumerate}
After a preliminary session, in which participants are familiarized with the execution, each subject performed $20$ trials per intention. The experimenter visually monitored each trial to ensure exact compliance of these requirements. In order to homogenize the dataset, we completely removed trials judged imprecise. Thus, the final dataset includes $1098$ trial ($253$ for pouring, $262$ for passing, $300$ for drinking and $283$ for placing) and, for each of them, both 3D and video data have been collected.

\paragraph{3D kinematic data.} Near-infrared $100$ Hz VICON system was used to track the hand kinematics. Nine cameras were placed in the experimental room and each participant's right hand was outfitted with $20$ lightweight retro-reflective hemispheric markers. After data collection, each trial was individually inspected for correct marker identification and then run through a low-pass Butterworth filter with a 6 Hz cutoff. Globally, each trial is represented with a set of 3D points describing the trajectory covered by every single marker during execution phase. The $x,y,z$ marker coordinates only consider the reach-to-grasp phase, where the following movement is totally discarded. Indeed, the acquisition of each trial is automatically ruled by a thresholding of the wrist velocity $v(t)$ at time $t,$ acquired by the corresponding marker. Being $\varepsilon = 20$ mm/s, at the first instant $t_0$ when $v(t_0) > \varepsilon,$ the acquisition starts and it is stopped at time $t_f,$ when the wrist velocity $v(t_f) < \varepsilon.$ 

\begin{figure}[t!]
	\centering
	\includegraphics[width=\columnwidth,keepaspectratio]{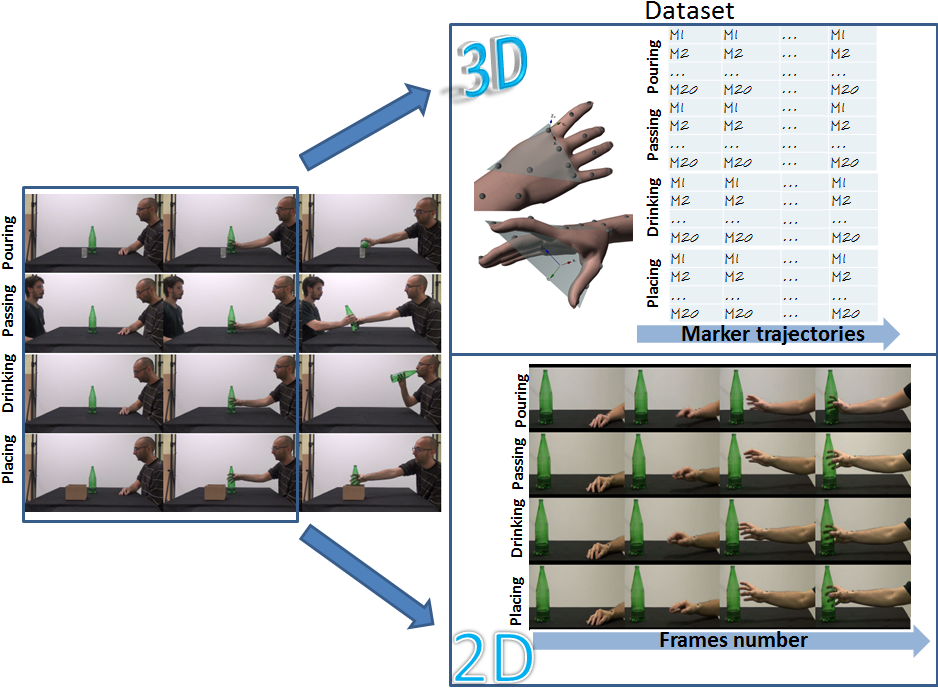}%
	\caption{The dataset. On the left we have the entire visible pouring, passing, drinking and placing development. On the top right, we have 3D VICON data acquisition, on bottom right, video sequences in which camera shoots only the arm and the bottle. In both cases, the acquisition stop at the grasping moment.}	
\end{figure}
\vspace{-.4 cm}

\paragraph{2D video sequences.} Movements were also filmed from a lateral viewpoint using a fixed digital video camera (Sony Handycam 3-D) placed at about $120$ cm from hand start position. The view angle is directed perpendicularly to the agent's midline, in order to ensure that the hand and the bottle were fully visible from the beginning up to the end of the movement. It is worth noting that the video camera was positioned in a way that neither the box (placing), nor the glass (pouring), nor the co-experimenter (passing) were filmed. Adobe Premiere Pro CS6 was used to edit the video in .mp4 format with disabled audio, 25 fps and  $1280 \times 800$ pixel resolution. In order to format video sequences in an identical way to 3D data, each video clip was cut off at the exact moment when the bottle is grasped, discarding everything happening afterwards. To better understanding how demanding the task is, note that the actual acquired video sequences encoding the grasping last for about one fourth of the future action we want to predict. Consequently all the sequences result about 30 frames long (some sequences are available in the Supplementary Material).

In all the experiments reported in this paper, either dealing with 3D or 2D data, we consider all the possible pairwise comparisons between intentions and the all-class one. We select one-subject-out testing procedure. That is, we compute seventeen accuracies, training our system on all the subjects except the one we are testing, then we averaged all the accuracies to get the final classification results. 

\section{Human performance in IfM}\label{sez:uomo}

As a preliminary analysis to check how human beings can predict intentions, we tested the human capabilities on Pouring
vs. Placing and Pouring vs. Drinking throughout the following experimental apparatus. We asked each of 36 participants to
watch 400 videos of reach-to-grasp movements and predict if it was finalized either to pour some water or pass the bottle. We
balanced the videos from each class (50 for Pouring and 50 for Placing, with 4 repetitions of each video). The experiment
starts showing the complete execution of the reach-to-grasp and its conclusion (Pouring or Placing) in a wide zoom where
the glass or the box, respectively, were visible. Then, we narrow the field of view, discarding everything except the arm,
the table and the bottle, and we show only the reach-to-grasp movement. After 8 demo trials in which the future intention
was revealed, we randomly shuffled the 400 videos and tested all the participants, registering their guess. Averaging all the
human accuracies in the Pouring vs. Placing test, we get 68\% of accuracy. Afterward, we move to the second test (Pouring vs.
Drinking) and we repeated the same procedure. In Pouring vs. Drinking, accuracy decreases to 58\% (-10\%). These results
are statistically significant and suggest that human observers are able to exploit kinematics to predict intentions. Globally,
although the human brain can read in a grasping some motion pattern which anticipates four different intentions, the computer
vision methods explained in the paper are more valuable and outperforming human predictive ability.

\section{3D kinematic analysis}\label{sez:3D}

Recent social cognitive work \cite{top} argued that intentions become ``visible'' in the apparent motion flow and understanding other's intentions cannot be divorced from the discrimination of essential kinematics. Thus, we exploited 3D kinematic features (KF) for IfM. Following \cite{Carpinella:2011}, we computed \emph{wrist velocity}, the module of the velocity of the wrist marker, \emph{wrist height}, the $z$-component of the wrist marker, \emph{wrist horizontal trajectory} defined as the $x$-component of the wrist marker and \emph{grip aperture}, \ie the distance thumb-index tips markers. Such features were referred to the reference system of the motion capture system, $F_{\rm global}$ \cite{Carpinella:2011}. A better characterization of the dynamics can be provided using a local reference system centered on the hand, $F_{\rm local}$ \cite{Asuini:2015}. In this way, we computed relative $x,y,z$ coordinates of thumb, index, thumb-index plane and the radius-phalanx. These variables provide the information about either the adduction/abduction movement of the thumb and index fingers or the  rotation of the hand dorsum. Thus, they ensure robustness towards finger flexion/extension or wrist rotation that can vary significantly from one trial to another \cite{Asuini:2015}. The 4 features from $F_{\rm global}$ and the 12 from $F_{\rm local}$ gives a total amount of 16 kinematic features. Acquisition time $[t_0,t_f]$ (see Section \ref{sez:dataset}) is scaled into $[0,1]$ and data are sub-sampled with step 0.01. Consequently, for each of our 16 kinematic features, we have $100$ equispaced values describing the evolution of such features during the reach-to-grasp movement. 

In Table \ref{tab:3D}, we report the classification results, using a linear support vector machine (SVM), considering $F_{\rm local}$ and $F_{\rm global}$ \cite{Carpinella:2011} individually, then concatenating all the features in $F_{\rm k}$ \cite{Asuini:2015}. In the three cases, we used a feature vector of 1200, 400 and 1600 components respectively. As expected, when we combine $F_{\rm local}$ and $F_{\rm global}$, the performance generally improves. Our results show that KF are able to obtain classification performances with a substantial improvement over the random guess level: using $F_{\rm k}$, an average +26,25\% improvement over the random guess level on the pairwise comparisons and +20,13\% on the all-class case. Relying on the kinematic interpretability of KF, we conclude that the actual dynamic of the grasping encodes some motion patterns which go beyond the fulfilment of the action itself and can concretely anticipate the underlying intention.
\begin{table}[t!]
	\centering
	\begin{tabular}{|c|c|c|c|}
		& \multicolumn{3}{|c|}{Linear SVM fed with KF}  \\
		& $F_{\rm local}$      & $F_{\rm global}$     & $F_{\rm k}$      \\ \hline\hline
		Pouring vs. Placing & 79,70 & 86,10 & 84,32  \\
		Pouring vs. Drinking & 72,15 & 70,36 & 76,48  \\
		Pouring vs. Passing & 76,55 & 67,39 & {\bf 82,81 } \\
		Passing vs. Drinking & 63,10 & 68,05 & 70,75  \\
		Passing vs. Placing & 62,60 & 64,38 & 69,44  \\
		Drinking vs. Placing & 64,40 & 71,41 & 73,72  \\\hline
		All-class    & 45,08 & 48,01 & 55,13 
	\end{tabular}
	\caption{Results from 3D data. For the kinematic features (KF), a linear $C=10$ SVM is fed with $F_{\rm local}$ and $F_{\rm global}$ groups of features, as well as with their combination $F_{\rm k}$.}
	\label{tab:3D}
	\vspace{-.2cm}
\end{table}


\section{Video-based analysis}\label{sez:2D}

To investigate the affordability of IfM we take advantage of some of the most effective spatio-temporal techniques \cite{Laptev:2005, Wang:2011}. We compare STIP \cite{Laptev:2005} and dense trajectories \cite{Wang:2011}, which will be shortened in Section \ref{sez:short}. In Section \ref{sez:perc} we perform an analysis considering only fragments of video frames. Finally, in Section \ref{sez:disc}, a discussion of the achieved results is reported.

\subsection{Local spatio-temporal features for IfM}\label{ss:pipe}
\begin{figure*}[h!tbp]
	\RawFloats\TopFloatBoxes
	\begin{floatrow}
		\capbtabbox{
			\centering
			\begin{tabular} {l|c|c|c}
				Comparisons	 & $\scriptstyle \mathcal{S} = 600$  & $\scriptstyle \mathcal{S} = 1000$ & $\scriptstyle \mathcal{S} = 2000$ \\ \hline
				{\scriptsize Pouring vs. Placing} & 79,69 & {\bf 81,22} & 79,41  \\ 
				{\scriptsize Pouring vs. Drinking} & {\bf  64,62} & 64,15 & 64,58  \\
				{\scriptsize Pouring vs. Passing} & 59,73 & 62,48 & {\bf  62,63}  \\
				{\scriptsize Passing vs Drinking} & {\bf  59,53} & 58,43 & 57,00 \\
				{\scriptsize Passing vs Placing} & 56,87 & 59,96 & {\bf 60,22}  \\
				{\scriptsize Drinking vs Placing} & 65,97 & {\bf 67,84} & 66,61  \\
				{\scriptsize All 4 intentions}  & 36,64 & {\bf 39,07} & 38,41  \\		
		\end{tabular}}{
			\caption{\scriptsize STIP classification percentages.}
			\label{sparse}}
		\capbtabbox{
			\centering
			\begin{tabular}{c|c|c|c|c|c}
				Comparisons & $\scriptstyle \mathcal{S} = 600$  & $\scriptstyle \mathcal{S} = 1000$ & $\scriptstyle \mathcal{S} = 2000$  & $\scriptstyle \mathcal{S} = 5000$ & $\scriptstyle \mathcal{S} = 10000$ \\ \hline
				{\scriptsize Pouring vs. Placing} & 82,65 & 82,76 & {\bf 83,18} & 82,45 & 82,99 \\ 
				{\scriptsize Pouring vs. Drinking} & 69,97 & 70,50 & {\bf 71,73} & 70,79 & 71,10 \\
				{\scriptsize Pouring vs. Passing} & 74,85 & 74,79 & 75,61 & 75,55 & {\bf 75,87} \\
				{\scriptsize Passing vs Drinking} & 67,20 & 66,62 & 68,37 & {\bf 68,95} & 67,98 \\
				{\scriptsize Passing vs Placing} & 68,00 & {\bf 68,46} & 67,16 & 68,00 & 68,25 \\
				{\scriptsize Drinking vs Placing} & 67,58 & 68,39 & 70,39 & 70,12 & {\bf 70,50} \\
				{\scriptsize All 4 intentions} & 46,08 & 46,76 & 47,24 & 47,18 & {\bf 47,35} \\ 		
		\end{tabular}}{
			\caption{\scriptsize Dense trajectories classification percentages $\mathcal{S} = 600,\dots,10000.$}
			\label{dense_gen}}
	\end{floatrow}
	
\end{figure*}


To perform action recognition, the class of approaches, named in \cite{arxiv} as local, extracts some interest points (IPs), detecting remarkable variations in both space and time, and associate each of them to a volume from which features are computed. Among the most effective methods, STIP \cite{Laptev:2005} and dense trajectories (DT) \cite{Wang:2011} use a \emph{sparse} or \emph{dense} approach, respectively, to extract IPs. For STIP \cite{Laptev:2005}, a convolution with a Gaussian filter is used, while for DT \cite{Wang:2011}, at different scales, a dense grid of points is initialized and each of them is tracked using optical flow. Subsequently, spatio-temporal volumes are generated by stacking the $N \times N$ spatial neighbours centered in all the $L$ points in any tracked trajectory. The volume is warped in order to follow the related trajectory and it is subdivided in a $n_x \times n_y \times n_t$ grid of cuboids. 
For any channel of features, a dictionary of visual words is created: if $B$ denotes the vocabulary size, each video $v$ is represented by a set of histograms $H^i,$ where $i$ runs over the channels. We have $H^i = \{h^i_1,\dots,h^i_B\},$ where $h^i_b > 0$ for any $b=1,\dots,B$ and we normalized $\sum_b h_b^i = 1$ for any $i.$  For the classification, the following exponential $\chi^2$ kernel is adopted.
\begin{equation}\label{eq:ker}
K(v,\overline{v}) = \exp \left( - \dfrac{1}{2} \sum_i \dfrac{d(H^i,\overline{H}^i)}{A_i} \right),
\end{equation}
where $v$ and $\overline{v}$ are two arbitrary videos (represented by the histograms $H^i$ and $\overline{H}^i$, respectively), and 
\begin{equation}
d(H^i,\overline{H}^i) = \sum_{b=1}^B \dfrac{(h^i_b -\overline{h}^i_b)^2}{h^i_b +\overline{h}^i_b},
\end{equation}
is the $\chi^2$ distance, whose average value is $A_i$ for channel $i.$ 

In our work, we used only two channels (HOG and HOF) and the kernel \eqref{eq:ker} fed a support vector machine with default parameter $C=10$ in a one-subject-out testing procedure\footnote{We compute seventeen accuracies training our systems on all the subjects except the one we are testing; then we average all the accuracies to get the final classification result.}.

\begin{table}[t!]
	\centering
	\begin{tabular}{c|c|c}
		Comparisons	& $L=15$    & $L=5$     \\ \hline
		{\scriptsize Pouring vs. Placing} & 82.76 & {\bf 86.99} \\
		{\scriptsize Pouring vs. Drinking} & 70.50  & {\bf 75.73} \\
		{\scriptsize Pouring vs. Passing} & {\bf 74.79} & 72.36 \\
		{\scriptsize Passing vs Drinking} & 66.62 & {\bf 67.13} \\
		{\scriptsize Passing vs Placing} & {\bf 68.46} & 67.58 \\
		{\scriptsize Drinking vs Placing} & 68.39 & {\bf 75.11} \\
		{\scriptsize All 4 intentions}  & 46.76 & {\bf 50.55} \\
	\end{tabular}
	
	\caption{ Trajectories shortening, $\mathcal{S}=1000$}
	\label{tab:L15oL5}

\end{table}

We experimentally compared the STIP \cite{Laptev:2005} and dense trajectories \cite{Wang:2011} on our dataset. For this purpose, we employ the available public codes\footnote{\url{http://lear.inrialpes.fr/software}} using the default parameters configuration for both representations. In the bag-of-feature encoding, different vocabulary sizes $\mathcal{S} = 600,1000,2000$ have been used for both STIP and dense trajectories. As the latter approach gives much more features (on average, 7588 per video against 428 for STIP), for the dense trajectories we also used $\mathcal{S} = 5000,10000$ visual words (see Table \ref{sparse} and \ref{dense_gen}). In addition, to limit the computational costs, we randomly sampled a subset of 900,000 HOG/HOF dense trajectory features to build the vocabulary.

In Tables \ref{sparse} and \ref{dense_gen} we present the classification results comparing STIP and dense trajectories. Despite we changed the size $\mathcal{S}$ of the dictionaries (one for HOG and HOF, separately), we registered no significant variations in accuracy. Globally, the dense trajectories outperforms STIP in all the comparisons. Even though the IPs from STIP are supposed to be most descriptive in each frame, dense trajectories give much more features which led to a big gain in performance. Indeed, comparing STIP and dense trajectories with $\mathcal{S}=1000$, the average improvement in the pairwise comparisons is +6.24\% and +7.69\% in all four comparisons. Thus, from now on, we will focus on dense trajectories.

\begin{figure*}[]
	\RawFloats\TopFloatBoxes
	\begin{floatrow}
		\capbfigbox{\hspace{-.4 cm}
			\centering
			\includegraphics[width = .3\textwidth, keepaspectratio]{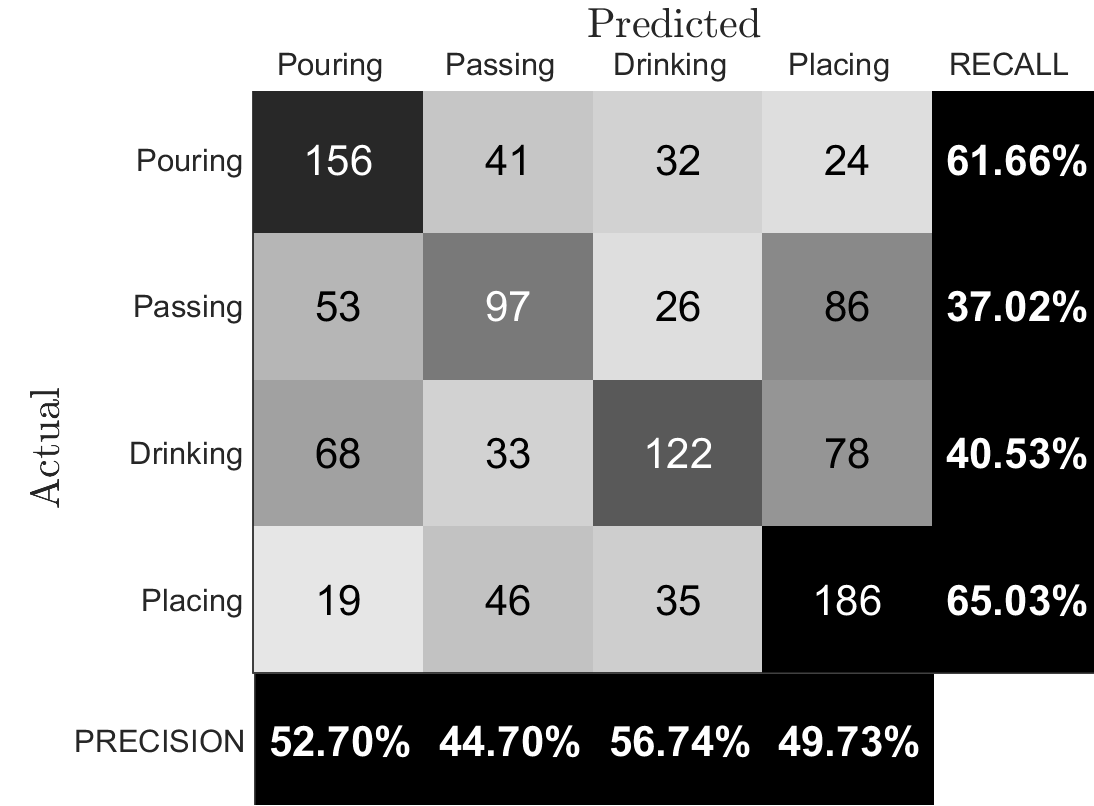}}
		{\caption{ Dense~trajectories~$L=5$~confusion~matrix.}
			\label{fig:conf}}
		\capbtabbox{
			\caption{Accuracy percentage for the snippet analysis.}
			\label{tab:perc}}{\hspace{-.2 cm}
			\centering
			\begin{tabular}{l|c|c|c|c|c|c|c}
				Comparisons & 40\%  & 50\%  & 60\%  & 70\%  & 80\%  & 90\% & 100\% \\\hline
				{\scriptsize Pouring vs. Placing} & 56,46 & 63,42 & 70,03 & 78,72 & 83,03 & 84,91 & {\bf 86.99}\\
				{\scriptsize Pouring vs. Drinking} & 52,99 & 60,73 & 64,06 & 64,55 & 67,61 & 72,13 & {\bf 75.73}\\
				{\scriptsize Pouring vs. Passing}& 61,02 & 61,77 & 62,57 & 65,05 & 65,55 & 69,83 & {\bf 72.36}\\
				{\scriptsize Passing vs. Drinking} & 60,92 & 62,67 & 66,49 & 69,30 & 66,69 & 63,97 & {\bf 67.13}\\
				{\scriptsize Passing vs. Placing} & 55,97 & 55,22 & 59,81 & 63,38 & 64,61 & 66,19 & {\bf 67.58}\\
				{\scriptsize Drinking vs. Placing} & 57,78 & 58,87 & 61,38 & 64,80 & 69,29 & 73,82 & {\bf 75.11}\\
				{\scriptsize All 4 intentions}   & 31,79 & 34,58 & 38,41 & 41,16 & 42,81 & 46,52 & {\bf 50.55}
		\end{tabular}}
	\end{floatrow}

\end{figure*}

\subsection{Dense trajectory shortening}\label{sez:short}

In order to tackle this extremely low variability between grasping motions, we tried to shorten the trajectory length changing the value of $L$ from $15$ to $5.$ With shorter trajectories, we tried to perform IfM analysis at a more atomic level. The rationale beside is that, since the clues that anticipate the intentions are almost invisible to the human eye (Section \ref{sez:uomo}), analysing shorter motion patterns may allow us to predict future intentions by a collection of subtle movements more easily discriminated. 
Operatively, we switched to $L=5$ and set $n_t = 1;$ we also fixed $\mathcal{S}=1000$ and, as done in Section \ref{ss:pipe}, we sampled a subset of 900,000 HOG/HOF. 
The risk of this approach is that features may lose discriminative power: in a 25 fps acquisition, our $L=5$ trajectories last $\frac{1}{5}$ seconds and, maybe, such time is too short to observe any useful motion cue. Fortunately, we found that this is not the case and, in fact, shorter trajectory features bring more evidence of the future intention. 
Actually, improving the beginning of the pipeline process results in a boost in the classification, showing that finer motion patterns are responsible of the actual intention.
In Table \ref{tab:L15oL5}, dense trajectory shortening is effective for IfM since provides +3.79\% accuracy improvement in the all-class comparison.

\subsection{Snippet analysis}\label{sez:perc}

Following the analysis in the previous section, we carried out a further investigation involving only small fragments of the video data. Shortening dense trajectories gives us the possibility of evaluating how much the movement is discriminant from the very beginning of the reach-to-grasp motion. In our case this is a further issue, since our dataset is made of very short videos, and considering only small portions we deal with only a bunch of frames for classification (e.g. in the 40\% analysis, only 6 frames for the shortest video). 
The snippet analysis has been performed in the following way. We considered the whole set of HOG an HOF descriptors and we built a global bag-of-features dictionary with $\mathcal{S}=1000$ words extracted from the 100\% of the frames. Then, for the analysis at a fixed rate, we keep only the descriptor computed over any spatio-temporal cuboid completely included in the considered portion of the video. This requirement explains why we skipped 10\%, 20\% and 30\%. Indeed, the shortest video lasts 16 frames and, thus, there are no spatio-temporal cuboids until we cover the 40\% of that particular reach-to-grasp movement.
In general, the accuracy increases as long as the percentage of the video considered grows (see Table \ref{tab:perc}): this means that bag-of-features histograms become more discriminative. Consequently, we can not find any portion of the reach-to-grasp execution that is useless for the prediction of intentions.

\subsection{Discussion}\label{sez:disc} Despite in Section \ref{ss:pipe} we showed that STIP \cite{Laptev:2005} are less effective than dense trajectories \cite{Wang:2011}, all the classification results in Table \ref{sparse} always overcomes the random chance discrimination, with the remarkable 81.22\% in Pouring vs. Placing task. Moving to the dense trajectory with $L=5$ (Table \ref{tab:L15oL5}), we have a high increase in performance on the pairwise comparisons (on average, +8.9\% improvement), and, again, the easiest one is Pouring vs. Placing  reaching 86.99\%. Instead, Drinking and Passing are the most problematic intentions to recognize (see Figure \ref{fig:conf}). 
In Table \ref{tab:L15oL5}, the all 4 intentions comparison leads to more impressive results: we indeed doubled the random chance level, scoring a 50.55\%, and thus improving STIP by +11.48\%. Consequently, we can conclude that, using computer vision, IfM motion problem is affordable in the sense that there exists some discriminative motion patterns that changes the execution of the reach-to-grasp movement, \emph{de facto} anticipating the future intention. Moreover, the snippet analysis shows remarkable results already using very few frames of the videos (Section \ref{sez:perc}). In addition, such results are extremely valuable if considering the improvement with respect to human beings (Section \ref{sez:uomo}).

Finally, one-subject-out testing procedure is suitable to devise an actual intention prediction system, dealing with human beings never seen before. However, it is much more demanding than a classical cross validation. For example, using $L=5$ dense trajectory, with a 10-fold cross validation procedure for testing, we reach 85.43\% in the all 4 comparison and 93.87\% (on average) in the pairwise comparisons.

\vspace{-.3 cm}

\section{Conclusions}\label{sez:con}

In this paper, we investigate the novel problem of Intention from Motion, consisting in the prediction of human intentions in a context-free setting, by only leveraging on kinematics. While proposing a novel multimodal dataset, we are able to show that the quantitative cognitive approach which rely on 3D motion capture data can be alternatively replaced with a video based paradigm where RGB optical videos are exploited. Despite the two data modalities are actually very different, we register the following common trends. First, random chance is always exceeded: therefore a context-free intention prediction is actually feasible. Second, in terms of performance, we register a similar level of classification accuracies while using classical kinematic features from cognitive literature and state-of-the-art video-based approaches in computer vision. We demonstrate that exploiting RGB videos for intention prediction is 1) easier to acquire and 2) equally reliable for a quantitative analysis as its motion capture counterpart. This finding opens to a joint interdisciplinary approach in taking advantage of computer vision methods while tackling cognitive problems such as the prediction of human intentions.

{\small
	\bibliographystyle{ieee}
	\bibliography{egbib}
}

\end{document}